\documentclass[journal=jacsat,manuscript=article]{achemso}

\usepackage{chemformula} 
\usepackage[T1]{fontenc} 
\usepackage{comment}
\usepackage{amsfonts}
\usepackage{amsmath}
\usepackage{commath}
\usepackage{graphicx}
\usepackage{caption}
\usepackage{subcaption}
\usepackage{lipsum}
\usepackage{wrapfig}
\usepackage{array}
\usepackage{url}
\usepackage[utf8]{inputenc}

\author{Jeonghee Jo}
\affiliation{Interdisciplinary Program in Bioinformatics, Seoul National University}

\author{Bumju Kwak}
\affiliation{Recommendation Team, Kakao Corporation}

\author{Byunghan Lee}
\affiliation{Department of Electronic and IT Media Engineering, Seoul National University of Science and Technology}

\author{Sungroh Yoon}
\affiliation{Interdisciplinary Program in Bioinformatics, Seoul National University}
\alsoaffiliation{Department of Electrical and Computer Engineering, Seoul National University}
\email{sryoon@snu.ac.kr}

\title[An \textsf{achemso} demo]
  {Flexible dual-branched message passing neural network for quantum mechanical property prediction with molecular conformation}

\abbreviations{IR,NMR,UV}

\begin{document}

\begin{abstract}

A molecule is a complex of heterogeneous components, and the spatial arrangements of these components determine the whole molecular properties and characteristics. With the advent of deep learning in computational chemistry, several studies have focused on how to predict molecular properties based on molecular configurations. Message passing neural network provides an effective framework for capturing molecular geometric features with the perspective of a molecule as a graph. However, most of these studies assumed that all heterogeneous molecular features, such as atomic charge, bond length, or other geometric features always contribute equivalently to the target prediction, regardless of the task type. 
In this study, we propose a dual-branched neural network for molecular property prediction based on message-passing framework. Our model learns heterogeneous molecular features with different scales, which are trained flexibly according to each prediction target. In addition, we introduce a discrete branch to learn single atom features without local aggregation, apart from message-passing steps. We verify that this novel structure can improve the model performance with faster convergence in most targets. The proposed model outperforms other recent models with sparser representations. Our experimental results indicate that in the chemical property prediction tasks, the diverse chemical nature of targets should be carefully considered for both model performance and generalizability. 

\end{abstract}

\begin{figure}
    \centering
    \includegraphics[scale=0.5]{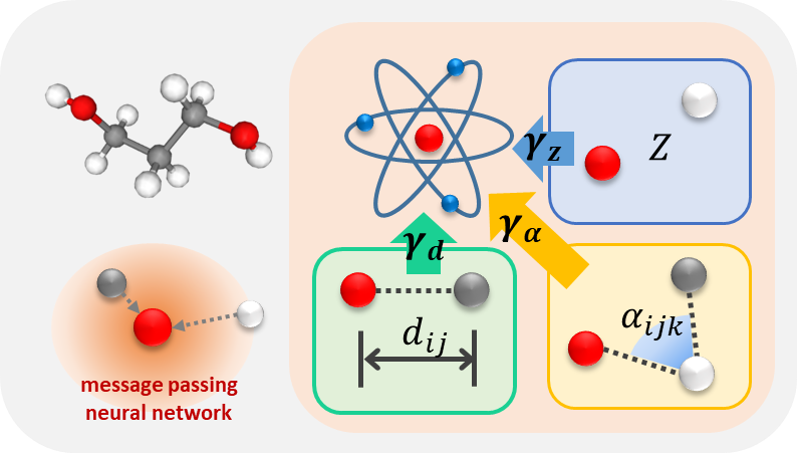}
    \caption{Abstract figure}
    \label{fig:abstract}
\end{figure}

\section{Introduction}

To design \textit{de novo} compounds, all possible sets should be explored throughout a combinatorial space called chemical compound space (CCS). Quantum mechanics (QM) functions as a guide that narrows down this search space based on the first principles of molecular dynamics. Density functional theory (DFT) is the standard method for analyzing a molecular electronic structure from its numerical wavefunctions in QM to predict molecular behaviors. DFT is crucial in modern quantum-based computational chemistry, which include quantitative structure activity relationship (QSAR) and quantitative structure property relationship (QSPR) analyses. In the real world, the reliability of these analyses has been reported to be significantly dependent on the quality and quantity of assessments \cite{ramakrishnan2014quantum}. However, the observations may be biased due to relatively insufficient experimental data or unexpected noises. 

In the past decade, machine learning (ML)-based methods have been developed, and have played a key role in various QM tasks. For example, support vector machine (SVM) \cite{gao2009accurate}, Gaussian processes \cite{bartok2010gaussian}, kernel ridge regression models \cite{rupp2012fast, snyder2012finding}, and other ML-based studies \cite{mills2011intramolecular} have been developed and applied to predict molecular properties or model molecular dynamics. Recently, with the advent of large-scale quantum mechanical databases, deep learning (DL)-based models have become crucial in quantum machine learning (QML) studies \cite{cova2019deep}. Deep neural network (DNN)-based methods have made it possible to model hidden complex relationships between molecular structure and their properties. 

Early DNN-based QM studies explored various tasks such as DFT prediction \cite{balabin2009neural}, a simple molecule dynamics simulation \cite{houlding2007polarizable}, atomization energy prediction \cite{hansen2013assessment}, or learning electronic properties \cite{montavon2013machine}. Some of these networks have the same number of input layer nodes (neurons) with the number of a given molecule, such that each neuron would get each atom representation \cite{lorenz2004representing, behler2007generalized}. These agrees with the basic scheme of the many-body problem in QM, in that the total potential energy of the system can be approximated with the sum operator of all local energies from particles consisting of. To be more concrete, a neuron of an input layer for a single atom can be considered as a one-body representation of a given system. 
Given a molecular geometry, a single atom can be represented with its atomic number and position, and multiple atoms can also be represented in the same with their interactions, using their mutual distances or angles. 
However, standard multi-layer perceptron (MLP)-based models cannot handle various sizes of chemicals \cite{behler2007generalized, bartok2010gaussian}. This limitation triggers a harmful effect on the model generalizability.

Graph neural networks (GNNs) can address these limitations by describing a molecule as a graph. GNNs utilize both node and edge attributes, which are naturally equivalent to atoms and atom-atom relationships present in a molecule. This graph-based molecular representation has become widely adopted in solving various molecule learning tasks. Several GNN-based models \cite{schutt2017schnet, chmiela2017machine, kondor2018n, unke2019physnet, wang2019molecule} adopted a localized convolution kernel to learn each component of a molecule. These models are called graph convolutional networks (GCNs). The spatial embedding based GCN model\footnote{Following from a GNN survey paper\cite{zhou2018graph, wu2020comprehensive} we denote \textit{a spatial-based} GCN as a type of GCN which uses a convolution filter on local context of a node. The MPNN\cite{gilmer2017neural} is also a type of spatial-based GCN.} trains each local attribute iteratively, and then aggregates these features to represent a molecular properties. 

The message-passing neural network (MPNN) framework \cite{gilmer2017neural} has taken these approaches one step further. Based on a graph convolution, a MPNN learns node features in two phases: \textit{message-passing} and \textit{update} phases. In the message-passing phase, the hidden representations of nodes, neighboring nodes, and their edges are aggregated to represent the locality of a given node, called a \textit{message}. Subsequently, the hidden features of these nodes are updated with the message and other features in the following update phase. The last layer, called a \textit{readout} phase, combines all of the features and outputs the result. In a MPNN framework, both the message aggregation and the readout phase use a sum function to gather local attributes and all atomic attributes, respectively. Based on this characteristic, the MPNN framework has several advantages over other architectures. Because a sum operation is order-invariant, the output representation is also invariant to the node ordering within a molecule. All MPNN models can be considered as a type of the spatial GCNs because in these models each atom feature is updated with its localized convolutional filters.

Since the late 2010s, several message passing-based DNN models have been developed, such as SchNet\cite{schutt2017schnet, schutt2018schnet}, GDML\cite{chmiela2017machine}, PhysNet\cite{unke2019physnet}, MegNet\cite{chen2019graph}, and DimeNet\cite{klicpera2020directional}. SchNet and the subsequent model architectures suggested \textit{a block} structure, which consists of a set of specific message passing functions. They also adopted a continuous-filter convolution to learn intramolecular interactions that take the form of continuous distribution. These models used atom types and atom-atom distances as inputs of the model. Some studies used more sophisticated functions to represent angles in a rotational equivariant approach\cite{anderson2019cormorant, miller2020relevance}.  Many of the MPNN-based models exhibited competitive performances in various QM research areas, even in the molecular property prediction on a non-equilibrium state. 

However, most of the previous works did not consider the natural heterogeneities among chemical properties. Although subtle, the target properties have originated from different chemical natures. For example, among the twelve targets from QM9, nine describe the types of given molecular energy, while other three do not correspond to. Conventional DL optimization can select the better parameters for describing all of the targets. However, it cannot determine that feature type that is more important for predicting each target. In this case, the optimal parameters in the given model architecture may be vary according to target types. Almost all previous models have no choice but to optimize parameters under the assumption that all types of features such as node features or edge features contribute equally to predicting all targets. This may limit the model generalizability and transferability because it ignores the diversity of chemical factors determining various properties. 

In addition, GNNs have suffered from differentiating nodes, especially with deep architecture, i.e., the \textit{over-smoothing} problem \cite{li2018deeper, zhou2018graph}. This problem emerges because an iterative aggregation of neighboring features is equivalent to a repeated mixing of local features, eventually making all node representations too similar to be differentiated from each other. This phenomenon reduces the model performance. Therefore, stacking more layers to a network cannot be the solution in GNN (MPNN) case without careful consideration.

To overcome all these limitations, we propose a novel GNN for molecular property prediction, A dual-branched Legendre message passing neural network (DL-MPNN) with simple and powerful scalability. We used two types of features: an atom type and a position in three-dimensional space. We calculated the distance between two atoms, and an angle between three atoms from atomic coordinates. In detail, we applied a modified Legendre rational functions for radial basis functions of atom-atom distances. For representation of angles between atoms, orthogonal Legengdre polynomials were used. Finally, we adopted trainable scalars to each type of features for model flexibility in multi-task QM learning. Our approaches exhibited superior performances to other recent models in quantum-mechanical benchmark tasks. In addition, we validate the effectiveness of the proposed method via ablation studies.

To the best of our knowledge, this study is the first attempt to automatically explore a relative feature importance for each target type without increasing computational cost. The introduction of the trainable scalars has several advantages over that of using an additional layer. First, the overall computational cost is not significantly increased because it only adopts scalar multiplications. 
Subsequently, it makes the model flexible for heterogeneous types of features in multi-target tasks.

In addition, we introduce an MLP-based novel-type block called a single-body block. These blocks are stacked to form a discrete branch in the model, and do not communicate with the message-passing branch, as they solely train only atom embeddings. The outcome of this MLP-based pathway is aggregated with that of the MPNN at the final stage. Accordingly, we obtain an atom feature that is not over-mixed by incoming messages during the training.

In summary, our contributions are as follows:
(1) We propose a novel dual-branched network for molecular property prediction, which comprises a message-passing for locally aggregated features and a fully-connected pathway for discrete atoms. (2) We introduce trainable scalar-valued parameters to enable the model enhance more important feature signals according to each label. (3) Our experimental results are better than those of most previous works in the public benchmarks with sparser representations. In addition, we verify that various quantum chemical factors contribute to differently according to target molecular properties.

\section{Related works}
Starting from DTNN \cite{schutt2017quantum}, several GNNs adopt the perspective of a molecule as a graph. Under this perspective, a molecular system is a combination of atoms and many-body interatomic interactions, that correspond to graph nodes and edges. MPNN \cite{gilmer2017neural} introduced the message concept, which is an aggregation of attaching edges and corresponding neighbor nodes. SchNet \cite{schutt2017schnet} is a multi-block network based on the message passing framework. The gradients flow atomwise, and the convolution process of atom-atom interaction features is in the interaction block. Subsequently, PhysNet \cite{unke2019physnet}, MegNet\cite{chen2019graph}, and MGCN\cite{lu2019molecular} extended the previous works and improved the performances based on the message-passing multi-block framework. 

More recent works introduced an angle information to describe the geometry of a molecule. With angle information, we can inspect up to three-body interactions. Spherical harmonics were used to represent an original angle to be rotationally equivariant \cite{poulenard2019effective, smidt2020euclidean}. Several studies \cite{kondor2018n, kondor2018clebsch, anderson2019cormorant, fuchs2020se} introduced a \textit{Clebsch-Gordan decomposition} to represent an angle comprising a linear combination of irreducible representations. Theoretically, it can be expanded to be arbitrary \textit{n-body networks}\cite{kondor2018n, anderson2019cormorant, fuchs2020se}, and most of the networks do not explore more than three-body interactions in limited computational resources. In summary, most of the recent GNNs on molecular learning used an atom type (one-body), a distance between an atom pair (two-body), and an angle between three atoms (three-body). As aforementioned, we also used the same input features in accordance with the previous GNNs, for fair evaluation.

\begin{figure}
  \centering
  
  \includegraphics[scale=0.6]{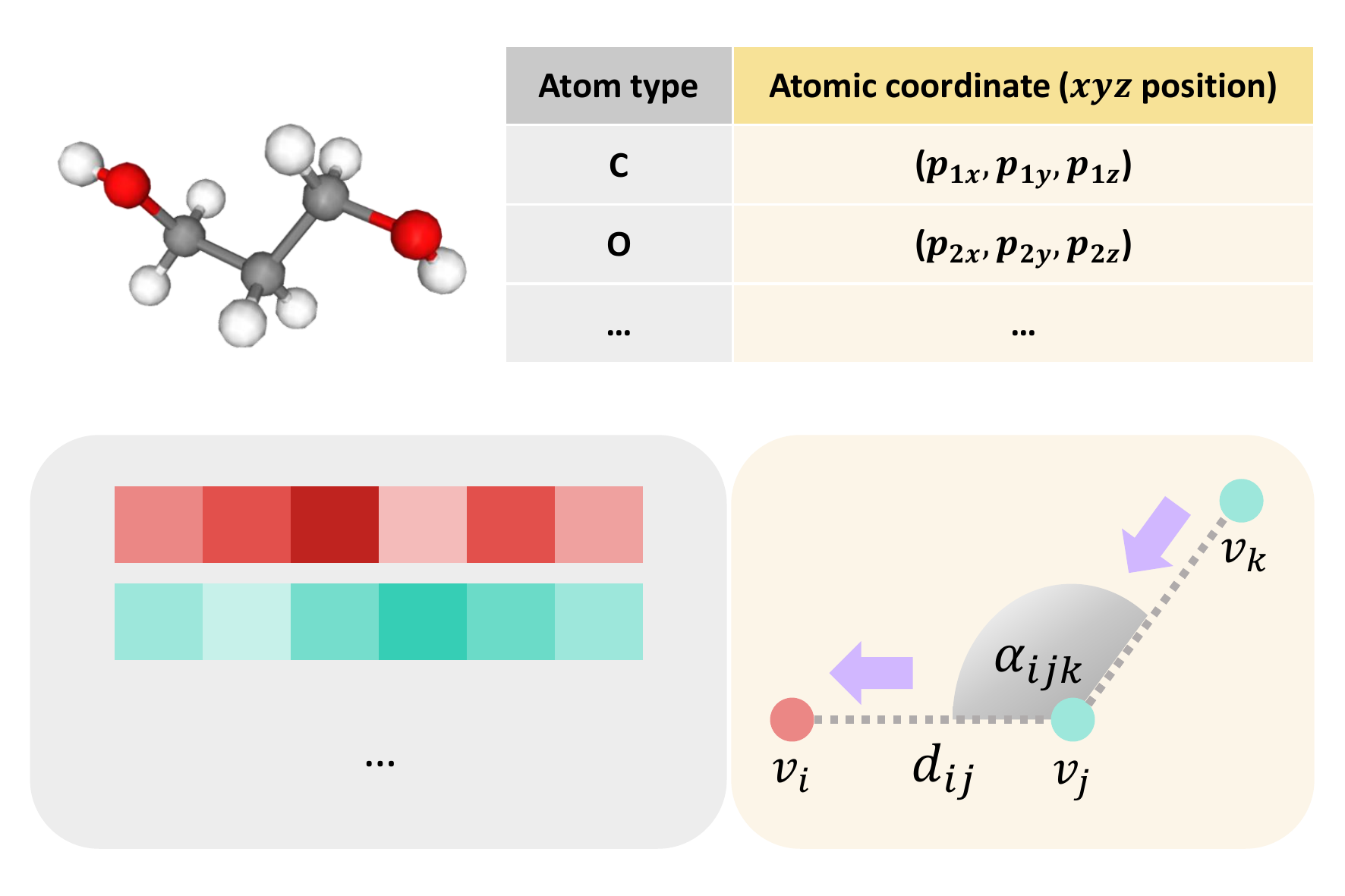}
  \caption{Data preprocessing. An atom type $Z$ and atomic coordinate $\mathrm{\mathbf{p}}$ were used in the model. We created trainable embeddings $X$ from $Z$, and calculated distances $d_{ij}$ and angles $\alpha_{ijk}$ from the coordinates $\mathrm{\mathbf{p_i}}$, $\mathrm{\mathbf{p_j}}$, and $\mathrm{\mathbf{p_k}}$.}
  \label{fig:preproc}
\end{figure}

\begin{figure}
  \centering
  
  \includegraphics[scale=0.55]{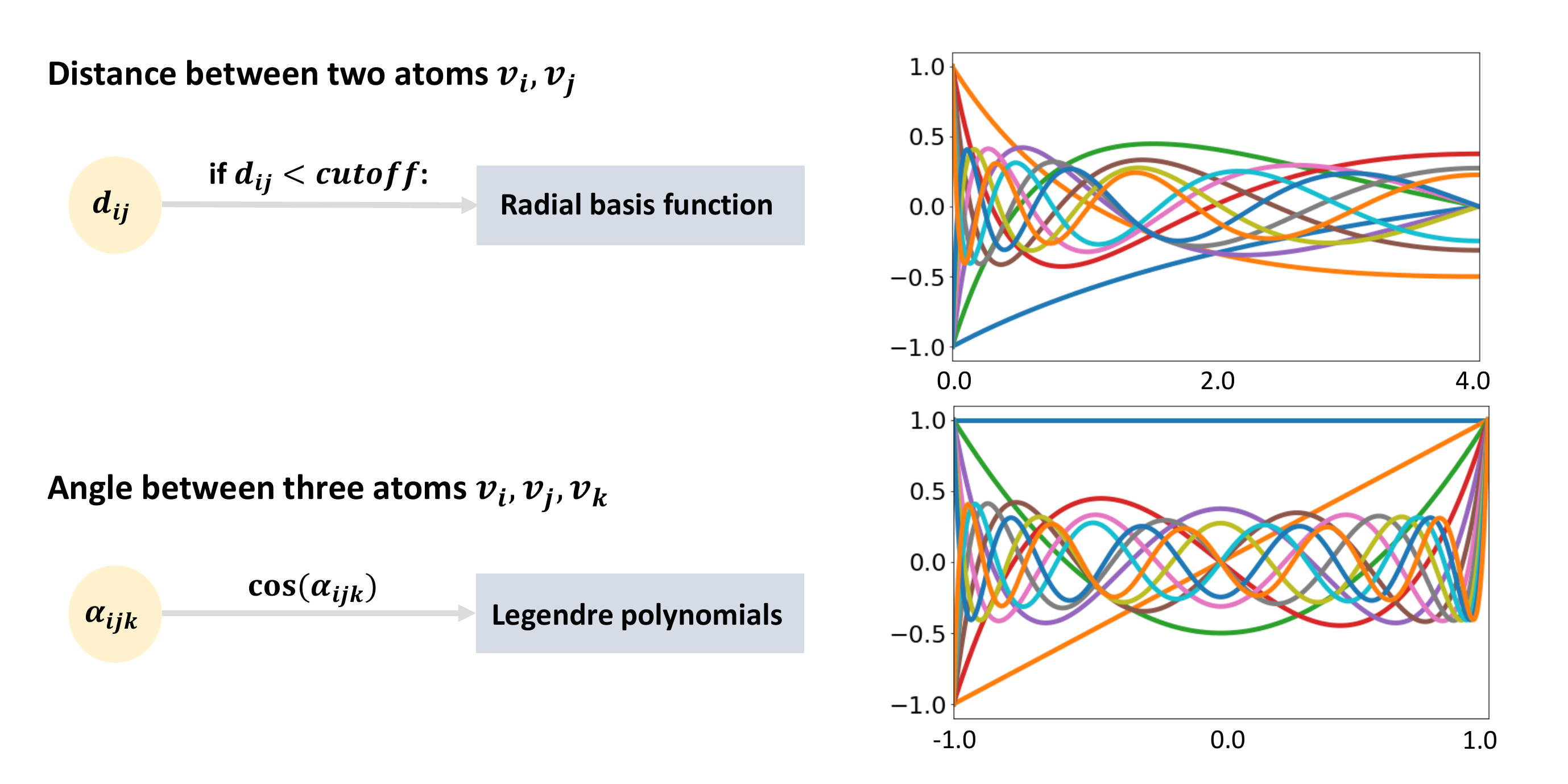}
  \caption{Basis functions for data preprocessing. For any pair of two atoms located closer than a given $cutoff$, we created an edge representation between two atoms regardless of molecular bond information. A scalar-valued distance is expanded as $n$-dimensional vector by radial basis functions. For radial basis functions, we used Legendre rational polynomials. Angle representations are depicted by any two edges sharing one atom. A cosine-valued angle is represented as the degree=$ 1, 2, ..., m $-th Legendre polynomials (top right). }
  \label{fig:basis}
\end{figure}

\begin{figure}
  \centering
  \includegraphics[scale=0.48]{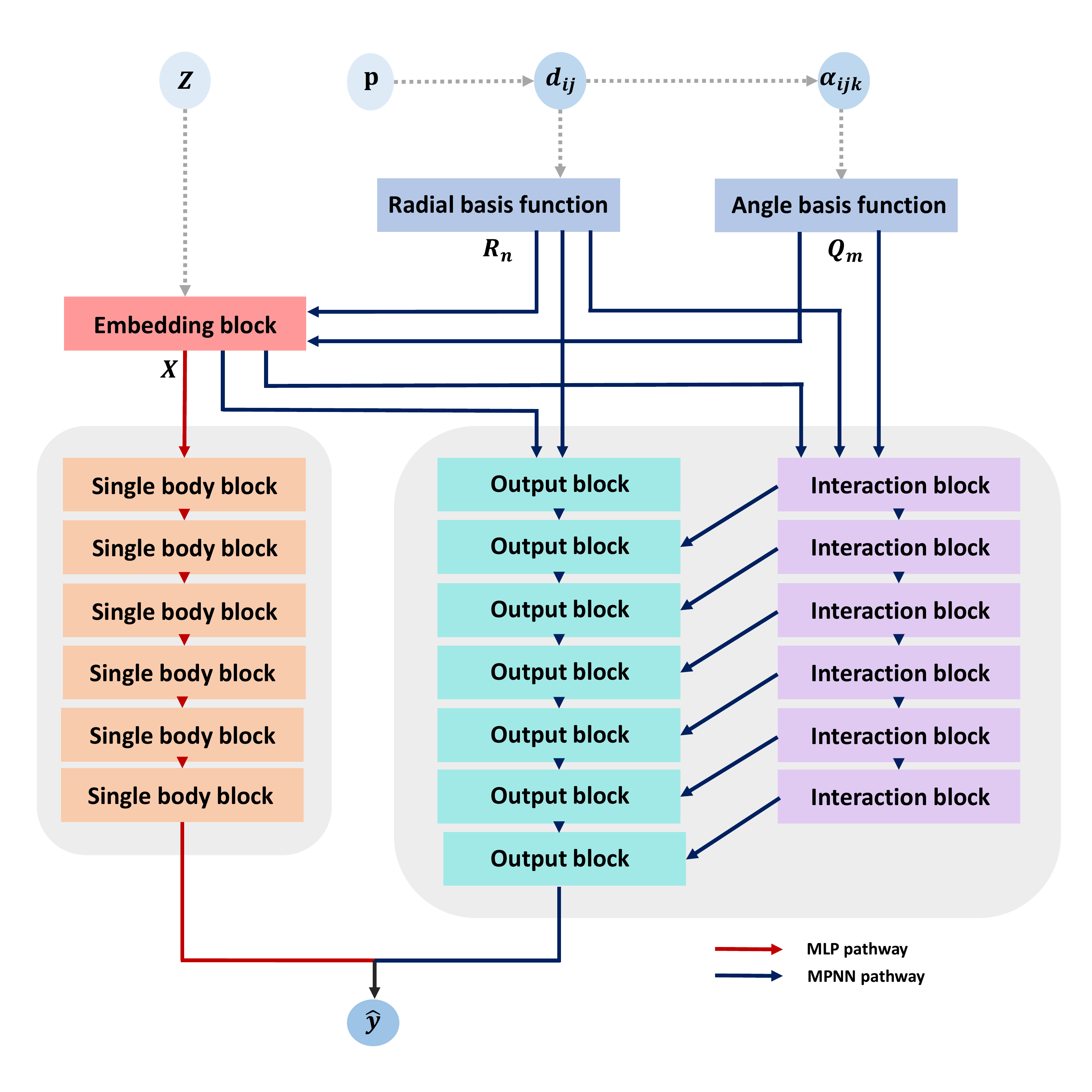}
  \caption{Model architecture. The network comprises two separate branches, MLP-based (red arrow) and MPNN-based (blue arrow) pathways. Atom types $Z$ are embedded as trainable matrices. Distances $d_{ij}$ and angles $\alpha_{ijk}$ are calculated from atomic coordinates. We calculated $R_{n}\in\mathbb{R}^{12}$ from $d_{ij}$. We also calculated $Q_{m}\in\mathbb{R}^{12}$ from $\alpha_{ijk}$. All blocks except the embedding block are stacked multiple times (not sharing weights). In this model, we stacked six, seven, and six blocks for the single body blocks, output blocks, and interaction blocks, respectively. For simplicity, skip connections are not shown.}
  \label{fig:archi}
\end{figure}

\begin{figure}
  \centering
  \includegraphics[scale=0.55]{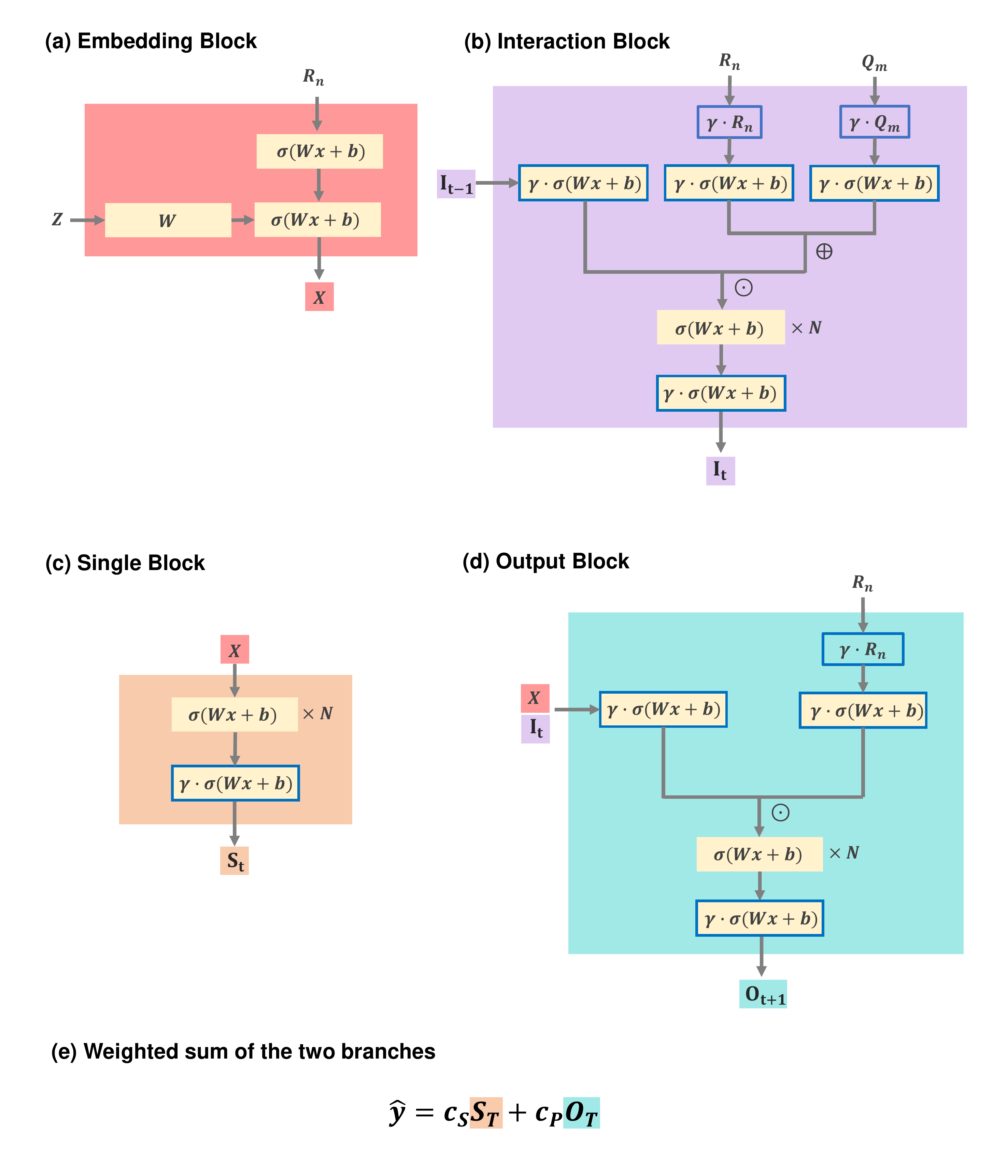}
  \caption{Block structures. (a) embedding block (left top). (b) interaction block (right top). (c) single body block (left bottom). (d) output block (right bottom). We denote the dense layer with the activation as $\sigma(Wx+b)$. We also denote the trainable scalars as $\gamma$. 
  All blue boxed items, including dense layers and features are \textit{$\gamma$-trainable}. Subscripts are omitted for simplicity. (e) weighted aggregation formula of two final outputs from the single block and output block pathways. $c_S$ and $c_P$ denote the trainable scalar-valued coefficients. }
  \label{fgr:blocks}
\end{figure}

\section{Methods}
Our neural network is dual-branched, comprising two different types of networks. One is the MPNN, a type of spatial-based GCNs, and the other is the MLP-based network. The details are described in the following sections.

\subsection{Preliminaries}
We briefly introduce the notations and the data used in this study. We describe a molecule as a set of different types of $i$ points $V=\{Z, P\}$ and $v_i=\{\mathrm{z_i}, \mathrm{p_i}\}$. $Z=\{z_i\}$ is a set of scalar-valued charges for each atom type. $ P=\{  \mathrm{\mathbf{p_i}}  \}, \text{where} \; \mathrm{\mathbf{p_i}}=(p_{ix}, p_{iy}, p_{iz})$ is a set of $xyz$ coordinates of atom points in the Euclidean space. We denote $v_{j, j \in N(i)}$ as a set of neighboring atoms of $v_i$.

\subsection{Data preprocessing}
We used two types of information between three neighboring atoms; atom types $z_i$ and coordinates $\mathrm{\mathbf{p_i}}$ of each atom in the molecules. We calculated distances $d_{ij}$ and angles $\alpha_{ijk}$ between atom coordinates $\mathrm{\mathbf{p_i}}, \mathrm{\mathbf{p_j}}, \text{ and } \mathrm{\mathbf{p_k}}$. Because coordinates are used to describe the whole molecular structure, a single position of an atom is meaningless. We initialized random trainable atom type embeddings $X$ from $Z$, and trained them. All $d_{ij}$ \text{and} $\alpha_{ijk}$ from the molecules were calculated before model training. More details are described in Figure \ref{fig:preproc}.

\subsection{Input embedding: Distance and angle representations}
\subsubsection{Distance representation: Radial basis function}
We used radial basis functions to expand a scalar-valued edge distance to a vector. In particular, we adopted sequences of Legendre rational polynomials, one of the continuous and bounded orthogonal polynomials. The Legendre rational function $R_n(x)$ and Legendre rational polynomial $P_n(x)$ are defined as: 

\begin{equation}
\label{eq:lrfunc}
    R_n(x)={ \frac{\sqrt{2}}{x+1} } P_n({ \frac{x-1}{x+1} })
\end{equation}

\begin{equation}
\label{eq:lpoly}
    P_n(x)={\frac{1}{2^n n!}}{\frac{d^n}{dx^n}}(x^2-1)^n
\end{equation}
where $n$ denotes the degree. The \textit{n-th order} Legendre rational polynomials are generated recursively as follows:
\begin{equation}
\label{eq:lrfrecur}
    R_{n+1}(x)={\frac{2n+1}{n+1}}{\frac{x-1}{x+1}}R_n(x)-{\frac{n}{n+1}}R_{n-1}(x)
    \; \text{ for } \: n \geq 1.
\end{equation}

We seleced the first \textit{n-th order} polynomials $R_1(x), R_2(x), ..., R_{n}(x)$, such that a scalar-valued distance $ d \in \mathbb{R} $ is embedded as an \textit{n}-dimensional vector $ \mathbf{d}\in \mathbb{R}^{n} $. The plot of functions degree of $1,2,...,12$ are described in top right side of Figure \ref{fig:basis}. We modified the equations to set the maximum value be $1.0$, to make the distribution of the bases similar to that of initial atom embedding distribution.

Recall that we did not use any molecular bond information. Following previous studies \cite{schutt2017schnet, schutt2018schnet, unke2019physnet, lu2019molecular, klicpera2020directional, anderson2019cormorant, miller2020relevance}, we set a single scalar cutoff $c$ and assumed that any atom pair $v_i$ and $v_j$ located closer than the cutoff can interact with each other, and vice versa, before training. In other words, we build edge representations $ \overrightarrow{e_{ij}}=(z_i, z_j, d_{ij}) $ between any two atoms close to each other. The cutoff value is analogous to kernel height or width in the conventional convolutional layer because it determines the size of the receptive field of the local feature map.

\subsubsection{Angle representation: Cosine basis function}
Similarly, we adopted another sequences of orthogonal polynomials to represent a scalar-valued angle $ \alpha_{ijk} $ as vectors: Legendre polynomials of the first kind. Legendre polynomials of the first kind are the orthogonal polynomials that exhibit several useful properties. They are the solutions to the Legendre differential equation\cite{anli2007some}. They are commonly used to describe various physical systems and dynamics. Their formula is expressed as:

\begin{equation}
\label{eq:defLp}
    Q_{n}(x)=2^{n}\sum_{k=0}^{n} x^{k} \binom{n}{k} \binom{{\frac{(n+k-1)}{2}}}{n}
\end{equation}

The polynomials of degree $n$ are orthogonal each other, such that:
\begin{equation}
\label{eq:ortho}
     \int_{-1}^{1} Q_{m}(x)Q_{n}(x) \,dx = 0 \;\;\;\; \text{if} \;\; n \neq m
\end{equation}

We calculated angles $ \alpha_{ijk} $ between any pair of two edges sharing one node, $ \overrightarrow{e_{ij}} $ and $ \overrightarrow{e_{jk}} $. We selected the first \textit{m-th order} polynomials $Q_1(x), Q_2(x), ..., Q_{m}(x)$, such that a scalar-valued angle $ \alpha_{ijk} \in \mathbb{R} $ is embedded as a \textit{m}-dimensional vector $ \mathbf{\alpha}\in \mathbb{R}^{m} $.  The scheme is presented on the left side of Figure \ref{fig:basis}. We calculated cosine values of each angle, and embed them with Legendre polynomials.

\subsubsection{Comparison of two different types of Legendre polynomials}
We briefly compare the two different Legendre polynomials. First, the sequences of Legendre rational functions $ \{P_k(x)\}_{k=1,...,n} $ are adopted as the encoding method for distances, which have different distributions over the distances. The standard deviations of the distributions of the distance encodings are higher at shorter distance values. Therefore, they can approximate the interatomic potentials, such that the atom pair exhibits stronger interatomic relationships if they are located closely with each other. In other words, the sequences of the Legendre rational functions $ \{ P_k(x) \} $ can represent make shorter distance values exhibit richer representations. 

Next, all Legendre polynomials of the first kind $ \{Q_k(x)\}_{k=1,...,m} $ are defined within the range of [-1, 1]. These functions are symmetric (even functions) or anti-symmetric (odd functions) at the zero point. Because the cosine function is an even function-type and ranges from -1 to 1, $ \{ Q_k(x) \} $ can cover the cosine angle of $ 0 $-$ 2\pi $, and is symmetric at the $ \pi $. In the three-dimensinoal Euclidean spaces, the angle of $ \theta \text{ } (0 \leq \theta \leq \pi) $ between two vectors is equal to the angle of $ (\pi - \theta) $. Therefore, Legendre polynomials of the first kind $ \{ Q_k(x) \} $ exhibit optimal properties for encoding the cosine of angles.


\subsection{Model architecture}
The overall model architecture is described in Figure \ref{fig:archi}. There are two discrete pathways in the network, which include the MLP-based pathway with single body blocks and MPNN-based pathway comprising output blocks and interaction blocks. First, an atom type $ Z $ is embedded by embedding block. The distances are calculated from atomic coordinates $ \mathrm{p} $, and if a distance $d_{ij}$ is less than the predefined $ cutoff $ value, then $d_{ij}$ is represented as an edge attribute of the molecule graph. The scalar $d_{ij}$ is expanded to a vector $\mathrm{d_{ij}}$ in the radial basis function $ \{P_k(x)\}_{k=1,...,n} $. The angles $\alpha_{ijk}$ are calculated from the edges. Then $ \alpha_{ijk} $ is encoded to an $ m $-dimensional vector before training via Legendre polynomials of the first kind $ \{Q_k(x)\}_{k=1,...,m} $. In this study, we set $m=12$.

In MPNN, the message passing and readout steps of our model can be summarized as:


\begin{align}
    m_i^{t+1} & = f_{m}^t(h_i^t, h_j^t, \overrightarrow{e_{ij}}, \alpha_{ijk}) \quad (t = 1,...,T) & \text{(message passing)} \nonumber \\
    h_i^{t+1} & = f_{u}^t(h_i^t, m_i^{t+1}) & \text{(update)} \label{eq:modelmp} \\
    \hat{y} & = f_{r}(v_i \in V \; | \; h^T) & \text{(readout)} \label{eq:modelreadout}
\end{align}
where $m$, $h$, $e$, and $\alpha$ denote the message of atom $v_i$ and its neighbors, single-atom representation of atom $v_i$, interaction between two neighboring atoms ($v_i$, $v_j$), and angle information between three atoms ($v_i$, $v_j$, $v_k$), respectively. $t$ is the time step or layer order in the model, and $\hat{y}$ represents the predicted target value. Both $f_{m}^t$ and $f_{u}^t$ represent the graph convolution, and $f_{r}$ is the sum operation.

In the proposed model, there are four types of blocks: the embedding block, output block, single body block, and interaction block. All blocks except the embedding block are sequentially stacked multiple times. Detailed explanations are presented in the next section. 

\subsection{Block details}

We adopted four types of blocks in the model. The embedding block $ \mathrm{E} $, the output blocks $ \mathrm{O_t} $, the single body blocks $ \mathrm{S_t} $, and the interaction blocks $ \mathrm{I_t} $, where $ t=0,...,(T-1)$ is the time step of the multiple sequential blocks. Atom type embeddings $ Z $ enter the embedding block $ E $ at the first step, and the embedded $ X $ moves to other three blocks at time step $ t=0 $. Distance embeddings $ \mathrm{d_{ij}} $ are used for all blocks $ \mathrm{E} $, $ \mathrm{O_t} $, and $ \mathrm{I_t} $, except the single body blocks $ \mathrm{S_t} $. Angle embeddings $ \mathrm{\alpha_{ij}} $ are soley used in $ \mathrm{I_t} $. All blocks are sequentially stacked with time steps $ t=0,...,(T-1)$ including some skip connections. The length of time steps $ T $ of each block may differ from other block types.

\subsubsection{embedding block}
In the embedding block $ \mathrm{E} $, the categorical atom types $ Z $ are represented as float-valued trainable features. $Z_i$, $Z_{j\in N(i)}$, and $\mathrm{\mathbf{d_{ij}}}$ from each atom pair $\overrightarrow{e_{ij}}$ are used to make atom embeddings $ X_i $. 

\subsubsection{output block}
In the output block $ \mathrm{O_t} $, the output of $ \mathrm{I_t} $ is trained with $R_{n}$, except the first block, which takes an embedded input $ \mathrm{X} $ from $ \mathrm{E} $. $\bigoplus$ and $\bigotimes$ denote the direct sum and direct product, respectively. All blue-boxed objects are $\gamma$-trainable. These blocks train the two-body representation.

\subsubsection{single body block}
In the single body block $ \mathrm{S_t} $, the output $X$ from $ \mathrm{E} $ enters, and are trained throughout multiple dense layers. The last of the layer is $\gamma$-trainable. No edge or angle representation is used. These blocks handle each-atom representations.

\subsubsection{interaction block}
In the interaction block $ \mathrm{I_t} $, and each output from the $ \mathrm{O_t} $ and $R_{n}$, $Q_{m}$ are applied. The blue boxes and the operators are also used in these blocks. In addition, these blocks train the three-body interaction.

\subsection{Trainable scalars}
In molecular property prediction tasks, various factors determine multiple properties with different contribution weights for each target. As aforementioned, we do not consider any additional layer or attentions preventing the model from the over-smoothing problem and becoming cost-intensive.

Instead, we propose a simple solution. As a solution, we introduced trainable scalars $\gamma$ to each layer. This makes the contributions of heterogeneous factors flexible, according to each target. Therefore, our model can focus on more important features of the target. Mathematical formulas are described below. 

\begin{equation}
\begin{split}
\label{eq:scalarlearn}
    h^{t+1} &= \sigma(Wh^t+b) \quad \text{(conventional dense layer)} \\
    h^{t+1} &= \gamma\ast\sigma(Wh^t+b) \quad \text{(with trainable scalars)}
\end{split}
\end{equation}

\begin{figure}
    \centering
    \includegraphics[width=0.65\textwidth]{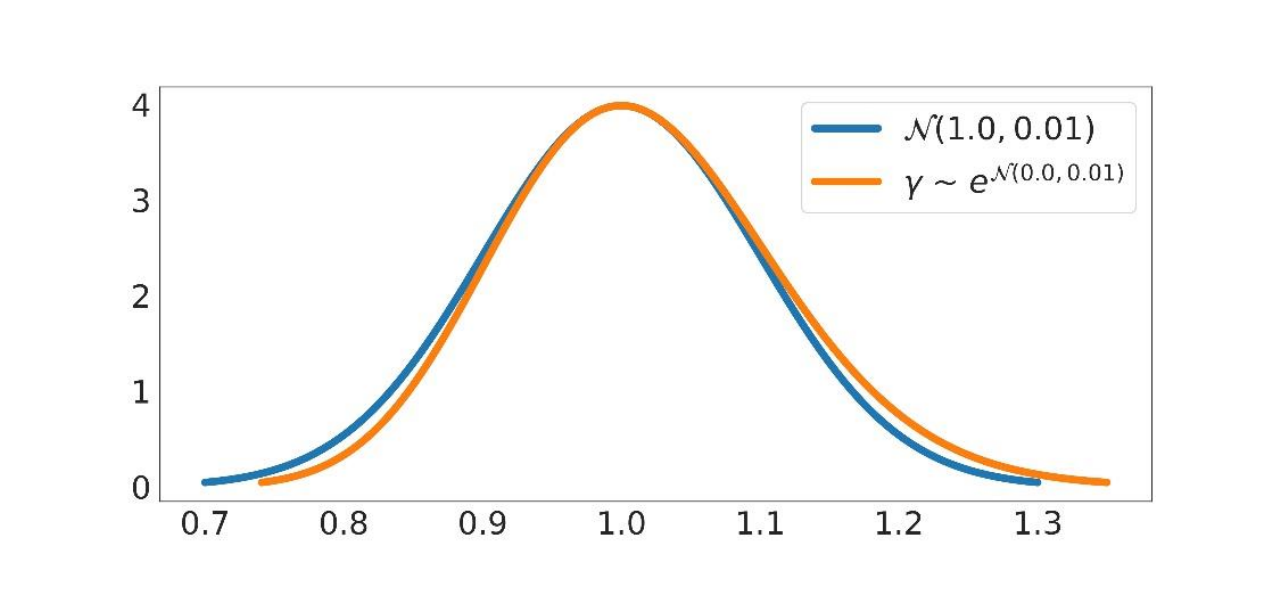}
    \caption{Gaussian distribution and the initialization of $\gamma$ distribution. Both distributions attain their maximum values at $x=1.0$}
    \label{fig:gdist}
\end{figure}

We introduced $ \gamma $ to some of the layers in the output blocks, single body blocks, and the interaction blocks. All $ \gamma $ values are trained independently of each other (subscripts were omitted for simplicity). We initialized $ \gamma $ values from the exponential function of random normal distribution $\mathcal{N}(0.0,\; 0.1^2)$ ($\mu = 0.0$, $\sigma = 0.1$), namely, $\gamma \sim e^{\mathcal{N}(0.0,\; 0.1^2)}$. This was because we observed that the slightly right-skewed distribution performed better than the normal distribution with $ \mu = 1.0 $, without skewness. This distribution is presented in Figure \ref{fig:gdist}.

\section{Training and evaluation}
\subsubsection{dataset} We used QM9\cite{ramakrishnan2014quantum} and molecular dynamics (MD) simulation datasets\cite{chmiela2017machine, chmiela2018towards} for the experiment. QM9 is the most popular quantum mechanics database created in 2014, which comprises 134k small molecules made up of five atoms, which include carbon, oxygen, hydrogen, nitrogen, and fluorine. Each molecule has twelve numerical properties, such that the benchmark consists of twelve regression tasks. The meaning of the targets is presented in Table \ref{tbl:qm9def}.

The MD simulation dataset is created for the energy prediction task from molecular conformational geometries. The simulation data are given as trajectories. The subtasks are divided according to each molecule type. The energy values are given as a scalar (kcal/mol), and forces are given as three-dimensional vectors (in $xyz$ format) of each atom. The energies can be predicted solely on the molecular geometry or by using the additional forces. 
We used the most recent subdatasets\cite{chmiela2018towards} of which the properties were created from CCSD(T) calculation, which is a more reliable method than conventional DFT method.

\subsubsection{implementation} The tensorflow package of version 2.2 was used to implement the proposed framework. To calculate Legendre polynomials, we utilized the scipy package version 1.5.2. 

For QM9 dataset, we trained the model at least 300 epochs for each target. We terminated the training when the validation mean absolute error (MAE) did not decrease for 30 epochs. Therefore, the overall training epochs are slightly different from each label. We randomly split the train, valid, and test dataset to 80\%, 10\%, and 10\% of whole dataset, respectively, in accordance with the guideline \cite{DBLP:journals/corr/WuRFGGPLP17}. The initial learning rate and decay rate were set to $10^{-3}$ and $0.96$ per 20 epochs, respectively. Adam\cite{kingma2014adam} was used as the optimizer, and the batch size was set to 24.

For the MD simulation dataset, most of the training configurations were the same with those of QM9 experiment. However, we modified the loss function because the energies were trained using both energies and forces, which differs from QM9. We obtained the predicted forces from the gradients of the atom positions following from the previous works \cite{unke2019physnet, klicpera2020directional}. We also followed the original guideline \cite{chmiela2018towards} for data splitting. Therefore, we used $ 1,000 $ samples to train all subtasks and the other $ 500 $ for test except Ethanol, and $ 1,000 $ to test of Ethanol. We also used the dropout\cite{srivastava2014dropout} rate as $0.33$ and set the decay rate as $0.9$ which makes the learning rate decrease faster than the QM9 training. 

\subsubsection{evaluation}
First, we evaluated the model performance with MAE, the standard metric of QM9 \cite{DBLP:journals/corr/WuRFGGPLP17}. We compared the performance of our model with the models of previous studies SchNet\cite{schutt2017schnet}, Cormorant\cite{anderson2019cormorant}, LieConv\cite{finzi2020generalizing}, DimeNet\cite{klicpera2020directional}, and DimeNet++\cite{klicpera_dimenetpp_2020}. We also analyzed our proposed methods in terms of the effect of $\gamma$ values, as well as the single body block of each target. We analyzed the three types of results, from the experiments with and without $\gamma$ or single body block, respectively. 

With the obtained $\gamma$ variables, we examined the contribution of trainable $\gamma$ values and the single body block to the prediction of each target. In particular, we observed that the single body block contributed to the model performance differently with each target type. We inferred that these differences can be explained by the different nature of properties. We discussed the relationship between some of the results and the chemical interpretations in the next section.

We compared the change in ratio of the average of all $\gamma$ values from the single block and that of the output blocks over epochs throughout the training. We extracted the $\gamma$ values and calculated the ratio of $\frac{\sum{(\gamma \text{ values from the single block})}}{ \sum{(\gamma \text{ values from the output block})}}$ every 30 epoch, until 240 epochs. After this point, the changes in the ratios varied negligibly, so we did not depict the ratios after that time point. Note that we did not compare the $\gamma$ values directly among different layers or targets. The magnitude of these values are determined by several complex sources: the original feature scale, layer weights, layer biases, activation functions, and uncontrollable random noises. Therefore, the $\gamma$ values themselves cannot be evaluated directly.

\begin{table}
  \caption{Description of the 12 targets of the QM9 dataset\cite{ramakrishnan2014quantum}}
  \label{tbl:qm9def}
  \begin{tabular}{ccl}
    \hline
    Property  & Unit & Description \\
    \hline
    $\mu$   & D  & Dipole moment \\
    $\alpha$ & $a_0^3$ & Isotropic polarizability \\
    $\epsilon_{\mathrm{HOMO}}$  & Ha & Energy of HOMO \\
    $\epsilon_{\mathrm{LUMO}}$ & Ha & Energy of LUMO \\
    $\epsilon_{\mathrm{gap}} $ & Ha & Gap($\epsilon_{\mathrm{LUMO}}$-$\epsilon_{\mathrm{HOMO}}$) \\
    <$R^2$> & $a_0^2$ & Electronic spatial extent \\
    zpve & Ha & Zero point vibrational energy \\
    $U_0$ & Ha & Internal energy at 0 K \\
    $U$ & Ha & Internal energy at 298.15 K \\
    $H$ & Ha & Enthalpy at 298.15 K \\
    $G$ & Ha & Free energy at 298.15 K \\
    $C_v$ & $\frac{cal}{molK}$ & Heat capacity at 298.15 K \\
    \hline
  \end{tabular}
\end{table}

\section{Result and Discussion}

\begin{table}
  \centering
  \caption{Mean absolute error on QM9 compared with previous works}
  \label{tbl:mae}
  \centering
  \begin{tabular}{llcccccc}
    \hline
    Target  & Unit & SchNet & Cormorant & LieConv & DimeNet & DimeNet++ & DL-MPNN \\
    \hline
    $\mu$ & D & 0.033 & 0.038 & 0.032 & 0.0286 & 0.0297 & \textbf{0.0256} \\
    $\alpha$ & ${bohr}^3$ & 0.235 & 0.085 & 0.084 & 0.0469 & \textbf{0.0435} & 0.0444 \\
    $\epsilon_{\mathrm{HOMO}}$ & eV & 0.041 & 0.034 & 0.030 & 0.0278 & 0.0246 & \textbf{0.0223} \\
    $\epsilon_{\mathrm{LUMO}}$ & eV & 0.034 &  0.038 & 0.025 & 0.0197 & 0.0195 & \textbf{0.0169} \\
    $\Delta_\epsilon$ & eV & 0.063 & 0.061 & 0.049 & 0.0348 & \textbf{0.0326} & 0.0391 \\
    <$R^2$> & ${bohr}^2$ & \textbf{0.073} & 0.961 & 0.800 & 0.331 & 0.331 & 0.414 \\
    $zpve$ & meV & 1.7 & 2.027 & 2.280 & 1.29 & \textbf{1.2} & \textbf{1.2} \\
    $U_0$ & eV & 0.014 & 0.022 & 0.019 & 0.00802 & 0.0063 & 0.0074 \\
    $U$ & eV & 0.019 & 0.021 & 0.019 & 0.00789 & \textbf{0.0063} & 0.0074 \\
    $H$ & eV & 0.014 & 0.021 & 0.024 & 0.00811 & 0.0065 & 0.0076 \\
    $G$ & eV & 0.014 & 0.020 & 0.022 & 0.00898 & \textbf{0.0076} & \textbf{0.0076} \\
    $C_v$ & $\frac{cal}{molK}$ & 0.033 & 0.026 & 0.038 & 0.0249 & 0.0249 & \textbf{0.023} \\
    \hline
  \end{tabular}
\end{table}

\begin{table}
    \caption{Mean absolute error on MD simulation compared with previous works}
    \label{tbl:md}
    \centering
    \begin{tabular}{llcccc}
        \hline
        Target & Train method & sGDML & SchNet & DimeNet & DL-MPNN \\
        \hline
         Aspirin & Forces & 0.68 & 1.35 & \textbf{0.499} & 0.590 \\
         Benzene & Forces & 0.06 & 0.31 & 0.187 & \textbf{0.053} \\
         Ethanol & Forces & 0.33 & 0.39 & 0.230 & \textbf{0.10} \\
         Malonaldehyde & Forces & 0.41 & 0.66 & 0.383 & \textbf{0.225} \\
         Toluene & Forces & 0.14 & 0.57 & 0.216 & \textbf{0.200} \\
         \hline
    \end{tabular}
\end{table}

\begin{figure}
    \centering
    \includegraphics[scale=0.6]{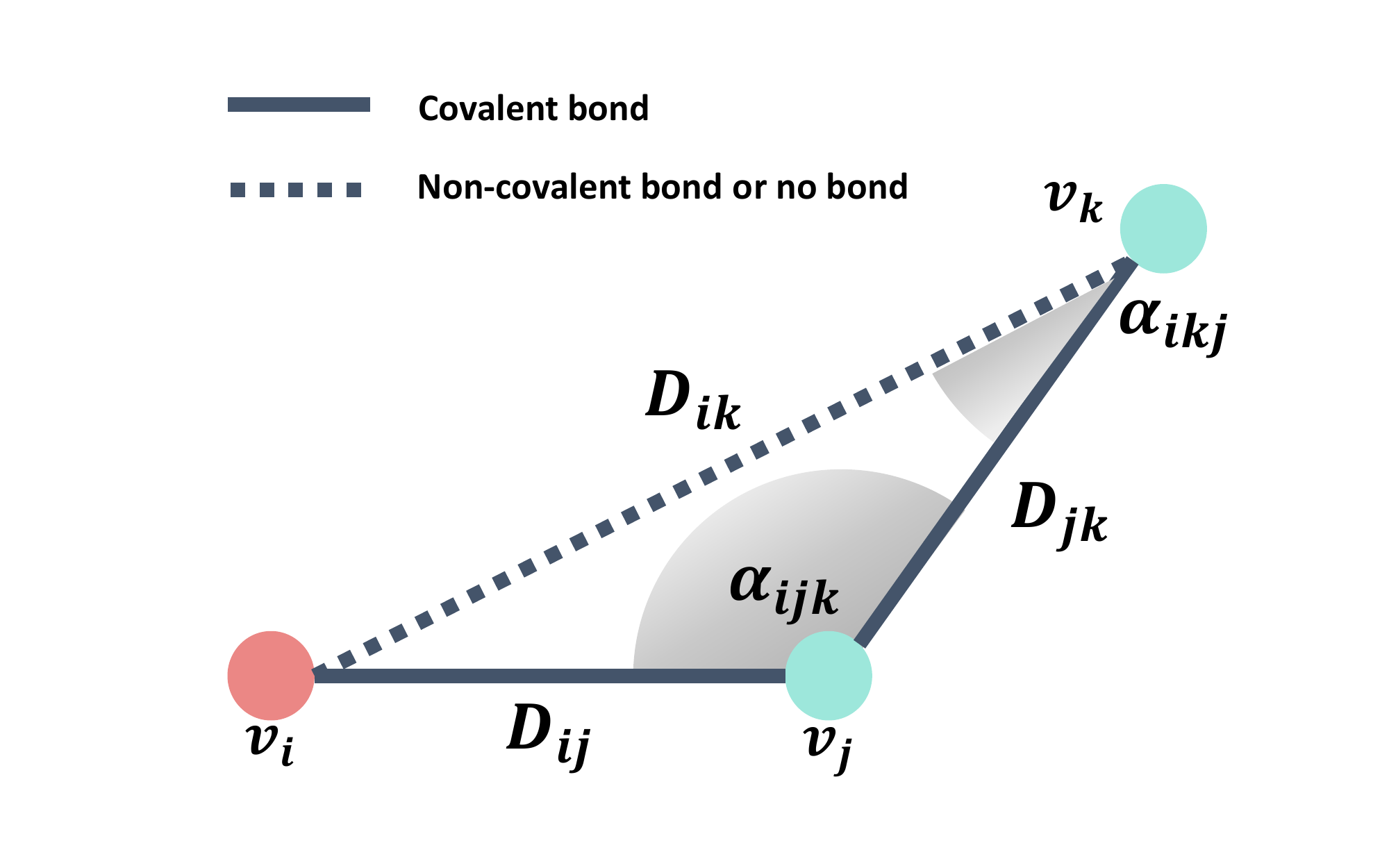}
    \caption{Example of bond and angle representation. All lengths of the covalent bond between two atoms of C, H, O, N, and F are less than $2.0$, so any two-hop neighbor via covalent bond $v_k$ from an atom $v_i$ will always be located inside the cutoff = $4.0$. Therefore, an atom $v_i$ can always capture two successive covalent bonds $(v_i, v_j)$ and $(v_j, v_k)$.}
    \label{fig:triangle}
\end{figure}

\subsection{Model Performance}
We compared the performance of our model performance with that of other MPNN-based models that also used atom types and locations as inputs \cite{schutt2017schnet, chen2019graph, anderson2019cormorant, klicpera2020directional, klicpera_dimenetpp_2020}. The MAEs for QM9 and thd MD simulation are described in Table \ref{tbl:mae} and for MD simulation in Table \ref{tbl:md}. For QM9, our model achieved advanced performances in six of the twelve targets. For the MD simulation, our model exhibited the best performance among other models in four of the five targets. 

In general, our model exhibited better performances on the targets in QM9, which are more related to molecular interactions such as dipole moment ($\mu$), molecular orbitals ($\epsilon_{\mathrm{HOMO}}$ and $\epsilon_{\mathrm{LUMO}}$), Gibbs free energy ($G$), and others. These properties are relevant in molecular reactions to external effects. However, predictions for other targets such as electronic spatial extent ($<R^2>$) and internal energies ($U_0$, $U$, $H$) were not superior to other models. 

We found that this may be related to the distribution of each target value. We observed that when the mean of the target value is closer to zero, and the standard deviation of the target value is smaller, the prediction performances are better. For example, the 95\% confidence intervals (CIs) of the distribution of HOMO and LUMO are $-6.5\pm{1.2}$ and $0.3\pm{1.3}$, respectively. In case of internal energies including $U$, $U_0$, and $H$, the 95\% of CI is $-76.5\pm{10.4}$, $-76.1\pm{10.3}$ and $-77.0\pm{10.5}$, respectively. These issue triggered by the diverse target distribution of QM9 has already been reported in the literature \cite{miller2020relevance}. In the future, we will analyse the effect of target distributions on the model convergence, in the future.

\subsection{Density of a molecule representation}
As mentioned before, the cutoff \textit{c} value determines the node connections in a molecular graph. With a high cutoff value, the molecular graph representation would be \textit{dense}. When the cutoff is low, the corresponding graph representation would be \textit{sparse}. 

The dense graphs generally have higher feasibilities in capturing the connectivity information in a graph than the sparse graphs. However, dense structures are exposed to higher risks of gathering excessive features even when there are less necessary information. Besides, in a MPNN, the messages of every node from its neighboring edges are mixed repeatedly throughout the network. If a graph is excessively dense, most messages become indistinguishable, because all atoms would refer all other neighbors. This may potentially increase the risk of over-smoothing problem\cite{li2018deeper}, prone to occur in deep GCNs.

It is challenging to determine the most appropriate cutoff value of the graph dataset; however, we found a clue from molecular conformations. We observed that the average distance between any atom pair in QM9 molecules is 3.27 angstrom (\AA). The atom pairs within 4.0 \AA\, which is the cutoff value used in this work account for 72\% of all the pairs in the QM9 dataset. Previous models SchNet\cite{schutt2017schnet}, MegNet\cite{chen2019graph}, and DimeNet\cite{klicpera2020directional}/DimeNet++\cite{klicpera_dimenetpp_2020} adopted 10.0, 5.0 and 5.0 as their cutoff values, respectively. We observed that 99.99\% and 89\% for cutoff values of $10.0$ and $5.0$ of the distances, respectively are represented in molecular graphs.

Note that the angle representation is defined with two different edges sharing one atom. This indicates that for the angle representation, the required number of computations increases linearly with the squared number of edges. If the number of nodes of a graph are $|V_G|$, then the upper bound of the number of representations including all nodes, edges, and angles of the graph is given as $|V_G|+{\frac{{|V_G||V_G-1|}}{2}}+{\frac{{|V_G||V_G-1||V_G-2|}}{6}}$ even if the message directions were not considered. We compared graph density values according to different cutoff values. If the edges are within a cutoff value of $4.0$, then the number of edge representations are reduced to the half ($(72\%)^2=52.8\%$) of the overall possible number of edges ${\frac{{|V_G||V_G-1|}}{2}}$. If the cutoff value is $5.0$, then the proportion of the number of the edge representations is increased to $80\%$ ($(89\%)^2=79.2\%$).

We demonstrate that the cutoff value of $4.0$ is sufficient in describing a molecular graph in QM9. As mentioned earlier, the molecules in QM9 comprise five atoms, C, H, O, N, and F. The lengths of the covalent bonds between any two atoms among these five atoms are always less than 2.0 \cite{allen1987tables}. Because the maximum length of any two successive covalent bondings is less than 4.0, the model can capture any two-hop neighboring atoms simultaneously, with a cutoff value of 4.0. This property makes it possible to identify all the angles between two covalent bonds. The detailed description is presented in Figure \ref{fig:triangle}. From these observations, we argue that the $4.0$ is the best choice as the cutoff value for efficient training and chemical nature. The same holds for MD simulation as well, because in this dataset, all the atoms are one of carbon, hydrogen, and oxygen (C, H, and O).

\section{Conclusion}
In this study, we developed a novel dual-branched network, message passing-based pathway for atom-atom interactions, and fully connected layers for single atom representations. We represented a molecule as a graph in three-dimensional space with atom types and atom positions. We embedded a scalar-valued atom-atom distance as a vector using Legendre Rational functions. Similarly, we embedded a scalar-valued angle as a vector using orthogonal Legendre polynomials. Both functions are complete orthogonal series and and have low computational costs. Our model exhibited remarkable performances in two quantum-chemical datasets. In addition, we proposed trainable scalar values, such that the proposed model can attend more significant features according to the various nature of the targets during the training. 
Furthermore, we adopted a smaller cutoff value than those used in previous MPNN models, which can help save the computational resources without loss of performance. Furthermore, we infer that this cutoff value is sufficiently long to identify the local structure of a molecule. In the future, we will conduct further analysis on the predictability of the model according to target distributions. Future works will be focused on other molecular property predictions or the predictions for more complex molecules.

\begin{acknowledgement}

\end{acknowledgement}

\bibliography{references}

\end{document}